\documentclass[11pt]{article}

\usepackage[final]{acl}

\usepackage{amsmath}
\newcommand{\quotes}[1]{``#1''}
\usepackage[table]{xcolor}
\usepackage{booktabs}
\usepackage{multirow}
\usepackage{amssymb}
\usepackage{pifont}

\newcommand{\xmark}{\ding{55}}
\usepackage{arydshln}
\usepackage{makecell}

\usepackage{times}
\usepackage{latexsym}

\usepackage[T1]{fontenc}

\usepackage[utf8]{inputenc}

\usepackage{microtype}

\usepackage{inconsolata}

\usepackage{graphicx}

\title{IMFD: End-to-end Multi-Face Forgery Detection through Instruction-based Large Vision-Language Models}

\author{
  Dasom Choi$^{1}$,
  Sangjun Moon$^{1}$,
  Hyeongchan Im$^{1}$,
  Jaeeon Park$^{3}$,
  $^{*}$Jingun Kwon$^{1}$, \\
  \textbf{Hidetaka Kamigaito}$^{2}$,
  \textbf{Taro Watanabe}$^{2}$,
  \textbf{and Manabu Okumura}$^{3}$ \\
  $^{1}$Chungnam National University \\
  $^{2}$Nara Institute of Science and Technology (NAIST),
  $^{3}$Institute of Science Tokyo \\
  {\ttfamily \{dasomchoi, sangjunmoon, hyeongchanim\}@o.cnu.ac.kr} \\
{\ttfamily jingun.kwon@cnu.ac.kr},
  {\ttfamily \{kamigaito.h, taro\}@is.naist.jp} \\
  {\ttfamily park.j.an@m.titech.ac.jp, oku@pi.titech.ac.jp}
  \\
}

\begin{document}
\maketitle
{\renewcommand{\thefootnote}{}
\footnotetext{%
$^{*}$ Corresponding author.\\
Our code is available at \url{https://github.com/somcandy08/IMFD}.
}}
\begin{abstract}
The rapid increase of deepfakes has raised significant concerns due to their spread on social media. Traditional multi-face forgery detectors crop and verify each face independently, ignoring background context and inter-face relationships, which often yields suboptimal performance. To overcome these limitations, we leverage instruction-based Large Vision–Language Models (LVLMs), which can interpret entire images and follow complex textual instructions. 
We propose a simple yet effective single-stage multi-face forgery detector, called IMFD (\underline{I}nstruction-based \underline{M}ulti-face \underline{F}orgery \underline{D}etector), which is trained end-to-end to jointly localize faces and predict per-face forgery labels. Rather than treating face box prediction only as a joint objective, IMFD explicitly integrates predicted face bounding boxes into the instruction as visual cues that enhance instruction grounding and forgery detection.
To support the training and evaluation of IMFD, we convert existing multi-face forgery datasets into an instruction-based format. Experimental results and analyses show that IMFD improves multi-face forgery detection by integrating face bounding boxes into the instruction, and consistently outperforms various state-of-the-art methods.
\end{abstract}

\section{Introduction}
\label{sec:intro}
The rapid advancement of diffusion-based generative models has enabled the synthesis of highly realistic images, audio, and videos from simple text instructions~\cite{dhariwal2021diffusionmodelsbeatgans}. While beneficial in industries such as film and advertising~\cite{xu2024mpdsmoviepostersdataset,wang2025generateecommerceproductbackground}, these models have also been increasingly abused to produce harmful content, specifically deepfake faces, undermining trust in digital media~\cite{social,279994}. Therefore, effective and reliable face forgery detection methods are needed.

Traditional approaches typically frame face forgery detection as binary classification and fall into three groups: methods using biological clues such as eye blinking, head poses, and skin texture~\cite{Liu_2020_CVPR}, signal-level artifacts to capture synthesis boundaries~\cite{he2021spectrumdetectingdeepfakesresynthesis}, and data-driven methods that train neural networks on labeled datasets~\cite{10.1109/TMM.2023.3313503}. Despite strong results in controlled settings, most methods adopt a two-stage pipeline: first cropping each face and then classifying each cropped face independently. This ignores global background context and inter-face relationships, leading to error propagation in scenes with many faces at varying scales and poses~\cite{9711250}.

To address these limitations, the OpenForensics dataset was introduced, enabling end-to-end multi-face forgery detection models~\cite{9711250}. To better distinguish real and forged faces, incorporating bi-grained contrastive learning into the detection backbone has been studied~\cite{zhang2024comicsendtoendbigrainedcontrastive}. However, these detectors remain object-centric and often miss subtle semantic manipulations. Recently, Large Vision–Language Models (LVLMs) combining large-scale vision encoders and language models have shown strong cross-modal reasoning~\cite{openai2024gpt4technicalreport}. They process whole images and complex prompts, reasoning over both global context and fine-grained details. 

Motivated by these advancements, we propose an instruction-based multi-face forgery detector, named \textbf{IMFD}. IMFD takes an image and a textual instruction as input, and is trained end-to-end with a joint objective that integrates two subtasks: instruction grounding with forgery detection and face bounding-box prediction. This single-stage framework teaches the model both where to look and what decision to make for each face by injecting predicted face coordinates into the instruction as visual cues. To improve robustness, we incorporate task-relevant details~\cite{misra-etal-2020-exploring}, such as the number of people, image resolution, and predicted face coordinates, directly into the instruction to emphasize information critical for detecting forged faces without external resources. For analysis, we also compare IMFD with a two-stage setting that uses ground-truth face coordinates. To support the training and evaluation of IMFD, we transform existing multi-face forgery benchmarks into an instruction-based format.

We conduct experiments on the large-scale OpenForensics dataset~\cite{9711250} and perform in-depth analyses to validate our method. Experimental results show that IMFD consistently outperforms various state-of-the-art baselines. 
The code and instruction-formatted dataset are publicly available at
\url{https://github.com/somcandy08/IMFD}.

\begin{figure}[t]
    \centering
    \includegraphics[width=1.0\linewidth]{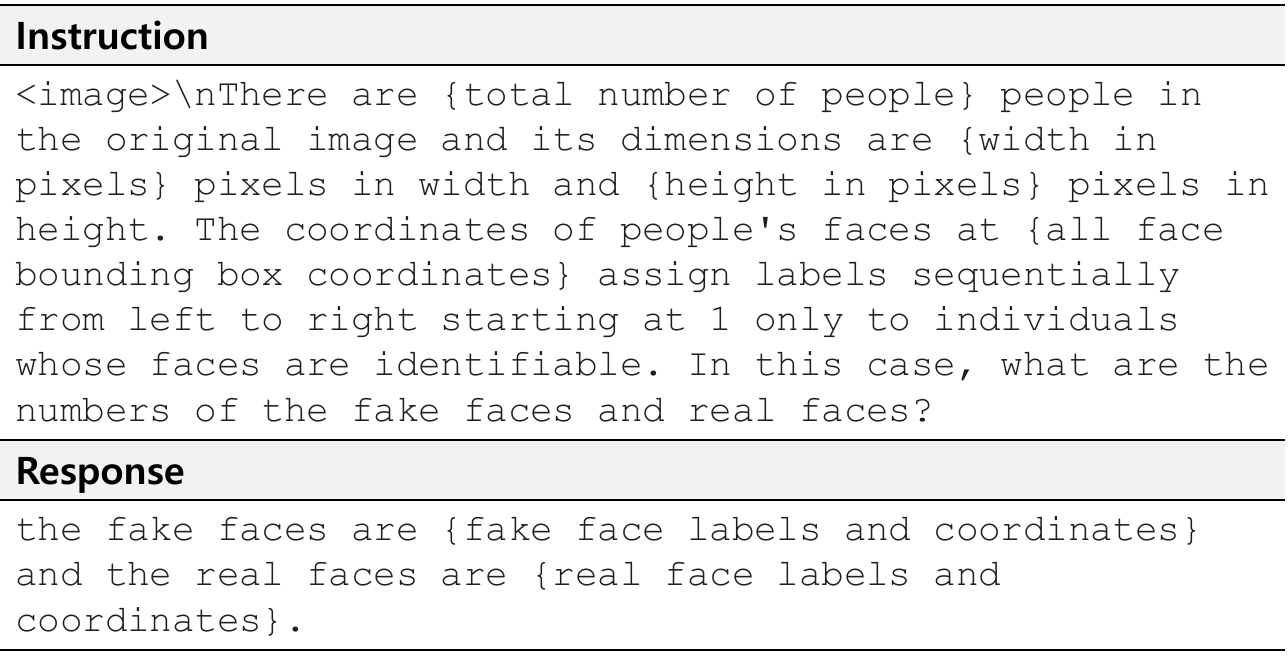}
    \caption{Instruction format for multi-face forgery detection using an ordered labeling output schema. \quotes{$<$image$>$} is a placeholder for the given image.}
    \label{fig:instruction_format}
\end{figure}

\section{Instruction-based Forgery Dataset}
Since no instruction-based datasets exist, we convert existing multi-face forgery datasets into an instruction-based format to train and evaluate IMFD in an instruction-based setting. Each sample pairs an image with an instruction that specifies how to index faces and how to format the output, ensuring consistent prompts and automatic evaluation. The LVLM is expected to identify and label any forged faces. To reduce typical LVLM errors even after fine-tuning~\cite{chen2023dress}, we augment the instruction with auxiliary details~\cite{misra-etal-2020-exploring}.

First, the total number of people is provided for the model to accurately verify face counts. Second, the input image resolution is included to help the model detect potential signs of forgery. Third, the face coordinate information explicitly directs the model's attention to specific facial regions to improve alignment between a given image and textual instructions. Figure~\ref{fig:instruction_format} shows the instruction format for multi-face forgery detection. With these details incorporated, the LVLM more effectively captures the correspondence between faces and the accompanying instruction.
We assign labels to faces from left to right, starting with the numeric number one based on the center coordinates of each face. We then embed these labels into the instruction format. The corresponding binary labels (real or fake) are inherited from the original dataset annotations. We include face bounding boxes in the output as well, enabling the model to better learn spatial grounding.

\section{Our End-to-end IMFD}
\begin{figure}[t]
    \centering
    \includegraphics[width=1.0\linewidth]{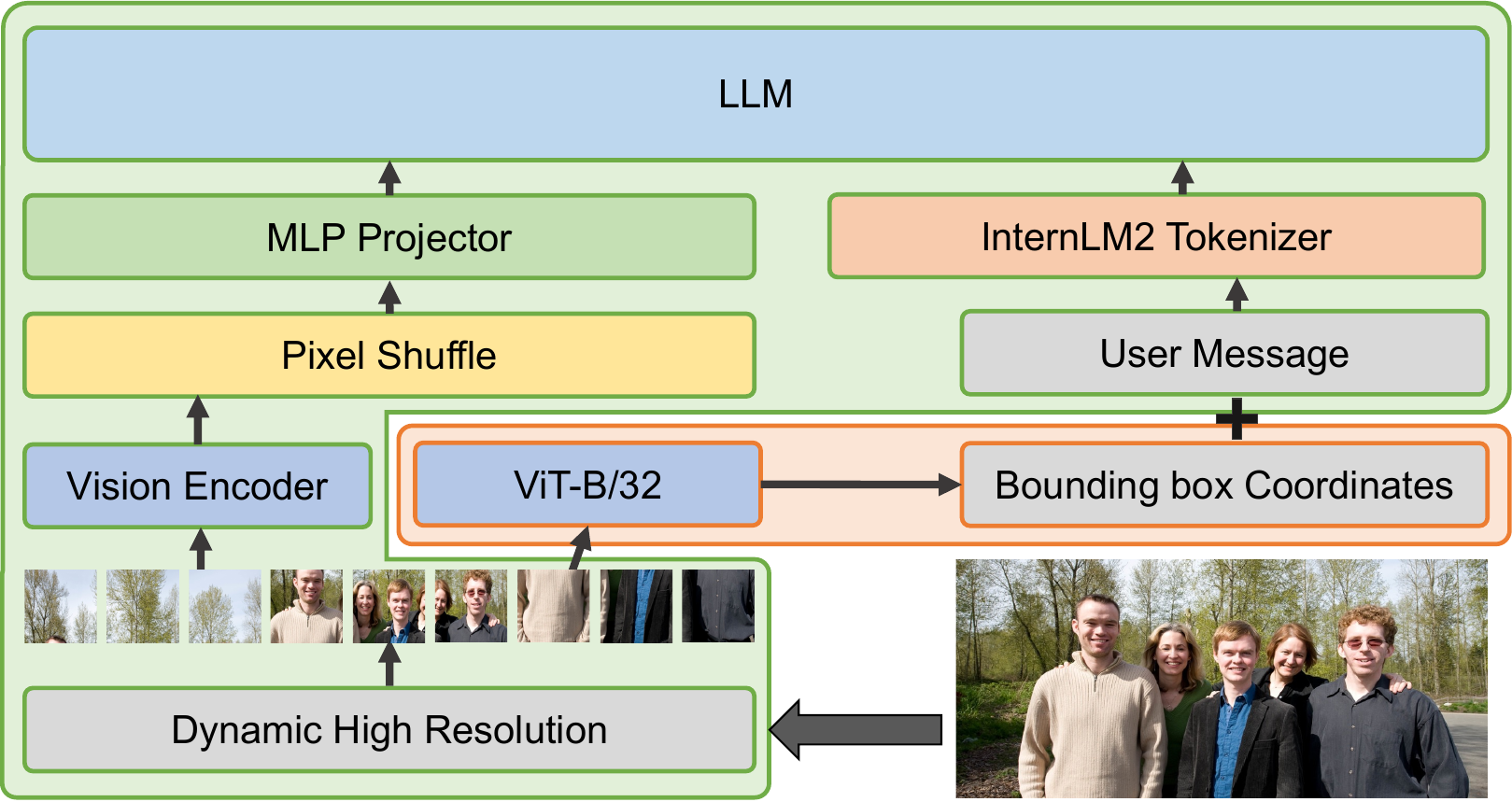}
    \caption{Overview of IMFD. The green block is a standard LVLM architecture such as InternVL. The red block predicts face boxes that are injected into the instruction.}
    \label{fig:e2e_main}
\end{figure}

We introduce our end-to-end IMFD. Figure~\ref{fig:e2e_main} shows an overview of the proposed model, which leverages LVLMs by jointly learning instruction grounding with forged face detection and bounding box prediction.
We define $\mathbf{X}_{img}$ as the input from the image modality, and $\mathbf{X}_{txt}=\{\mathrm{[img]}, w_1, w_2, ..., w_n\}$ as the input from the textual modality, where $\mathrm{[img]}$ is a special token that will later be replaced by the image embedding.

\noindent \textbf{Vision Encoder.}
Given $\mathbf{X}_{img}$, we obtain patch embeddings:

\begin{small}
\begin{align}
& X_0 \!= \![\mathbf{x}_{\mathrm{cls}} ; \mathrm{PatchEmbed}(X_{\mathrm{img}})] \!+ \!E_{\mathrm{pos}}, \\
& X'_l \!= \!\mathrm{MHA}(\mathrm{LN}(X_{l-1})) \!\!+ \!\!X_{l-1}, \!\hfill  (l = 1,..., L) \\
& X_l \!= \!\mathrm{MLP}_1(\mathrm{LN}(X'_l)) \!+\! X'_l,
\end{align}
\end{small}


where $\mathbf{x}_{\mathrm{cls}}$ is the special token, and $E_{\mathrm{pos}}$ denotes the positional embedding. LN, MHA, and MLP denote layer normalization, multi-head attention, and a feed-forward layer.

\noindent \textbf{Face Bounding Box Prediction.}
Face bounding boxes are predicted from a single text query, \quotes{A photo of face}, without explicit class prediction. Given an image $\mathbf{X}_{img}$ and anchor coordinates $Bboxes$, the procedure is as follows.

\noindent \textbf{Text–Anchor Similarity.}
A pre-trained CLIP ViT-B/32 encodes the query to a feature $F_{\mathrm{query}}$, which we L\textsubscript{2}-normalize~\cite{radford2021learningtransferablevisualmodels}.
For each anchor feature $F_{\mathrm{anchor},i}$, we compute the following:

{\small
\begin{flalign}
& S_i = F_{\mathrm{anchor},i} \cdot F_{\mathrm{query}}^{\top}, \quad
S = [S_1,\dots,S_N] \in \mathbb{R}^N, && \\
& F_{\mathrm{anchor},i} =
\frac{\mathrm{CLIP}_{\mathrm{img}}(c_i)}
{\lVert \mathrm{CLIP}_{\mathrm{img}}(c_i) \rVert_2},
\quad (i=1,\ldots,N) &&
\end{flalign}
}

where $F_{\mathrm{anchor}}$ is the normalized feature for the $i$-th anchor, computed by passing the anchor crop $c_i$ from the input image $\mathbf{X}_{img}$ through the pre-trained Vision Transformer CLIP image encoder architecture. Here, $N$ denotes the number of anchors. To obtain the anchor coordinates $c_i$, we first divide the image into a grid of rectangular regions and record each cell's coordinates as an anchor. For each grid cell, we generate additional anchor boxes by applying various width and height coefficients to scale the region around the cell’s center~\cite{radford2021learningtransferablevisualmodels}.

\noindent \textbf{Anchor Selection.}
Scores are min–max‑normalized and anchors whose scores exceed the skip threshold $\tau_{\mathrm{skip}}$ are retained: $\mathcal{I} = \{\, i \mid S_i \ge \tau_{\mathrm{skip}} \,\}.$
If no anchor satisfies the threshold (i.e., $\mathcal{I} = \varnothing$), we select the anchor with the highest score only if the image–text similarity $s_{\mathrm{img}} = F_{\mathrm{img}} \cdot F_{\mathrm{text}}^{\top}$ exceeds the true-positive threshold $\tau_{\mathrm{tp}}$, where $F_{\mathrm{img}}$ is obtained by passing $\mathbf{X}_{img}$ through the CLIP image encoder. If $s_{\mathrm{img}}$ falls below the false-positive threshold $\tau_{\mathrm{fp}}$, all boxes are discarded.

\noindent \textbf{Face Box Selection and Refinement.}
The selected anchors and their scores are refined using Weighted Boxes Fusion (WBF) to remove overlapping anchors:
$B_{pred} = \mathrm{WBF}(Bboxes[\mathcal{I}], S ;\ \tau_{\mathrm{iou}}),$ where $\tau_{\mathrm{iou}}$ is the IoU threshold used to determine which overlapping boxes are fused together.

\noindent \textbf{Integrating LLM and Vision Encoder.}
We map image features to the LLM embedding space and replace the \texttt{[img]} slot with the projected vector. Specifically,

\begin{small}
\begin{align}
F_{\mathrm{img}} &= \mathrm{MLP}_2(X_L[:,1:,:]), \\
X_0 &= \mathrm{Embed}(X_{\mathrm{txt}}) + E_{\mathrm{pos}}, \\
X_0[i] &= F_{\mathrm{img}}, \quad
Y_{\mathrm{pred}} = \mathrm{LLM}(X_0),
\end{align}
\end{small}

where $MLP_2$ projects the image embeddings to the same dimensionality as the word embedding space in the LLM. $Embed(X_{txt})$ denotes the token embeddings of the input text, and $i$ is the position of the special \texttt{[img]} token appended to the text input. The image representation $F_{img}$ replaces the embedding at this position, allowing the LLM to incorporate visual information during generation.

\noindent \textbf{Objective Function.}
We optimize the average of text and box losses:

{\small
\begin{align}
\mathcal{L}_{\mathrm{gen}} &= \mathrm{CrossEntropy}(Y_{\mathrm{pred}}, Y), \\
\mathcal{L}_{\mathrm{bbox}} &= \mathrm{L1Loss}(B_{\mathrm{pred}}, B), \\
\mathcal{L} &= (\mathcal{L}_{\mathrm{gen}} + \mathcal{L}_{\mathrm{bbox}})/2.
\end{align}
}
where $Y$ and $B$ denote the ground-truth text and boxes, respectively. 

During training, we use ground-truth face bounding-box coordinates. At test time, we use the predicted face bounding-box coordinates and derive the total number of people, which are then injected into the instruction. Image resolution is read directly from the input image. 

\begin{table*}[t]
\centering

\renewcommand{\arraystretch}{1.0}
\small
\begin{tabular}{lcccccc|ccc}
    \hline
    \rowcolor{gray!10}
    & \multicolumn{3}{c}{\textbf{ALL}}
    & \multicolumn{3}{c|}{\textbf{LARGE}}
    &\multicolumn{3}{c}{\textbf{ALL}} \\
    \rowcolor{gray!10}
    \textbf{Model}
    & \textbf{F1$_c$} & \textbf{F1$_a$} & \textbf{Acc}
    & \textbf{F1$_c$} & \textbf{F1$_a$} & \textbf{Acc}
    & \textbf{Lat.~$\downarrow$} & \textbf{Max Lat.~$\downarrow$} & \textbf{Thpt.~$\uparrow$}  \\
    \hline
    \multicolumn{10}{l}{\textbf{Two-Stage Methods}} \\
    \hline
    EfficientNet & 0.69 & 0.29 & 0.54 & 0.90 & 0.46 & 0.36 & \textbf{8.0} & \underline{363.7} & \textbf{128.8} \\
    Capsule      & \underline{0.70} & 0.29 & \underline{0.55} & \underline{0.90} & \underline{0.47} & \underline{0.56} & 14.9 & 1166.8 & 67.1 \\
    UCF          & 0.23 & 0.14 & 0.06 & 0.48 & 0.30 & 0.09 & 31.5 & 5394.2 & 31.9 \\
    F3Net        & 0.65 & \underline{0.31} & 0.35 & 0.81 & 0.45 & 0.44 & 43.8 & 939.1 & 22.8 \\
    \hdashline
    IMFD     & \textbf{0.98}$^\dagger$ & \textbf{0.98}$^\dagger$ & \textbf{0.94}$^\dagger$
                 & \textbf{0.97}$^\dagger$ & \textbf{0.98}$^\dagger$ & \textbf{0.87}$^\dagger$
                 & 162.8 & \textbf{263.7}$^\dagger$ & 6.3 \\
    \hline
    \multicolumn{10}{l}{\textbf{Single-Stage Methods}} \\
    \hline
    COMICS       & \underline{0.81} & \underline{0.74} & \underline{0.40} & \textbf{0.81} & \underline{0.74} & \underline{0.33} & \textbf{457.7} & \textbf{605.3} & \textbf{4.4} \\
    \hdashline
    IMFD     & \textbf{0.90}$^\dagger$ & \textbf{0.84}$^\dagger$ & \textbf{0.77}$^\dagger$
                 & 0.75 & \textbf{0.80}$^\dagger$ & \textbf{0.40}$^\dagger$
                 & 565.3 & 875.2 & 3.6 \\
    \hline
\end{tabular}
\caption{Comparison with SOTA baselines. Latency (Lat.) and maximum latency (Max Lat.) are reported in milliseconds, and throughput (Thpt.) is the number of images processed per second. $\dagger$ indicates that the improvement is significant (\textit{p}$<$0.01) compared with the underlined score (typically the best baseline), as determined by paired bootstrap resampling~\cite{koehn-sig}.
}
\label{tab:all_results}
\end{table*}

\section{Experiments}
\subsection{Experimental Settings}
\noindent \textbf{Datasets.}
We used OpenForensics, a large-scale multi-face forgery detection dataset~\cite{9711250}. 
The training, validation, and test-development datasets consist of 44,122, 7,308, and 18,895 pairs, respectively. For the test dataset used in the evaluation (ALL), the fake face ratio was 0.58. To investigate the multi-face forgery detection performance of the LVLM on images with a large number of faces, we also conducted evaluations on only 1,522 images (LARGE) that contain more than the average number of faces (2.63) in the test-development dataset.


\noindent \textbf{Implementation Details.}
We employed InternVL2-2B~\cite{chen2024internvl} as the backbone model and fine-tuned it with LoRA~\cite{hu2022lora}. Detailed settings are provided in Appendix~\ref{app:implementation_details}.

\noindent \textbf{Compared Models.}
Following the settings in~\cite{DeepfakeBench_YAN_NEURIPS2023,zhang2024comicsendtoendbigrainedcontrastive}, we compared our end-to-end \textbf{IMFD} with two-stage state-of-the-art (SOTA) baselines including Capsule~\cite{8682602}, UCF~\cite{yan2023ucfuncoveringcommonfeatures}, EfficientNet-B4~\cite{tan2020efficientnetrethinkingmodelscaling}, F3Net~\cite{qian2020thinkingfrequencyfaceforgery}, and the single-stage SOTA method, COMICS~\cite{zhang2024comicsendtoendbigrainedcontrastive}. All baselines were trained on the standard OpenForensics splits, whereas IMFD was trained on our instruction-based variant. We used ground-truth face bounding boxes for the two-stage setting, while predicted face bounding boxes were used for the single-stage setting. Additional upper-bound results obtained by injecting ground-truth face bounding boxes coordinates into the instruction, along with comparisons across instruction variants, are provided in Appendix~\ref{app:upper_bound_analysis}.

\noindent \textbf{Evaluation Metrics.}
For IMFD, we evaluate micro F1 (F$1_c$) for overall performance, macro F1 (F$1_a$) to balance rare labels, and exact matching (Acc), which counts an image as correct only when the set of predicted fake faces exactly matches the ground truth.
For two-stage models, we collected the independently detected results to calculate $F1_{c}$, $F1_{a}$, and $Acc$. For the single-stage COMICS model, to evaluate detection results as a classification problem, we assign a fake face when a predicted bounding box overlaps with a ground-truth box by an IoU threshold of 0.5 or higher. 

\begin{table}[t!]
\renewcommand{\arraystretch}{0.9}
    \centering
    \resizebox{1.0\columnwidth}{!}{
    \begin{tabular}{ccccccc}
        \toprule
        \rowcolor{gray!10}
        \textbf{Model} & \textbf{\# Peop.} & \textbf{Image Rel.} & \textbf{Coordi.} & \textbf{F1$_c$} & \textbf{F1$_a$} & \textbf{Acc} \\
        \midrule
        \multirow{6}{*}{IMFD}  
        & \xmark & \xmark & \xmark & 0.91 & 0.61 & 0.83 \\
        & \checkmark & \xmark & \xmark  & 0.94 & 0.81 & 0.87\\
        
        & \xmark & \checkmark & \xmark & 0.73 & 0.56 & 0.32 \\
        & \xmark & \xmark & \checkmark & 0.96 & 0.66 & 0.92\\
        
        & \xmark & \checkmark & \checkmark & \textbf{0.98} & 0.93 &  \textbf{0.96}\\
        & \checkmark & \checkmark & \checkmark &  \textbf{0.98} & \textbf{0.98} &  0.94\\
        \bottomrule
    \end{tabular}}
    \caption{Ablation studies on ALL regarding the effectiveness of integrating auxiliary details into instructions.}
    \label{tab:infpriming}
\end{table}

\noindent \textbf{Results.}
As shown by the results in Table~\ref{tab:all_results}, IMFD outperforms the other models in both two- and single-stage settings. This demonstrates the effectiveness of IMFD and the instruction-based dataset. While there is a trade-off between computational cost and detection performance, IMFD achieves significantly lower maximum latency compared to all two-stage methods in the two-stage setting. This is because, when a large number of faces are present in a single image, IMFD avoids the costly and sequential steps of face cropping and per-face detection required by traditional methods. However, the single-stage IMFD, which uses predicted face bounding boxes, shows a noticeable drop in performance compared to the two-stage IMFD that leverages ground-truth bounding boxes.

\begin{figure}[t]
    \centering
    \includegraphics[width=1.0\linewidth]{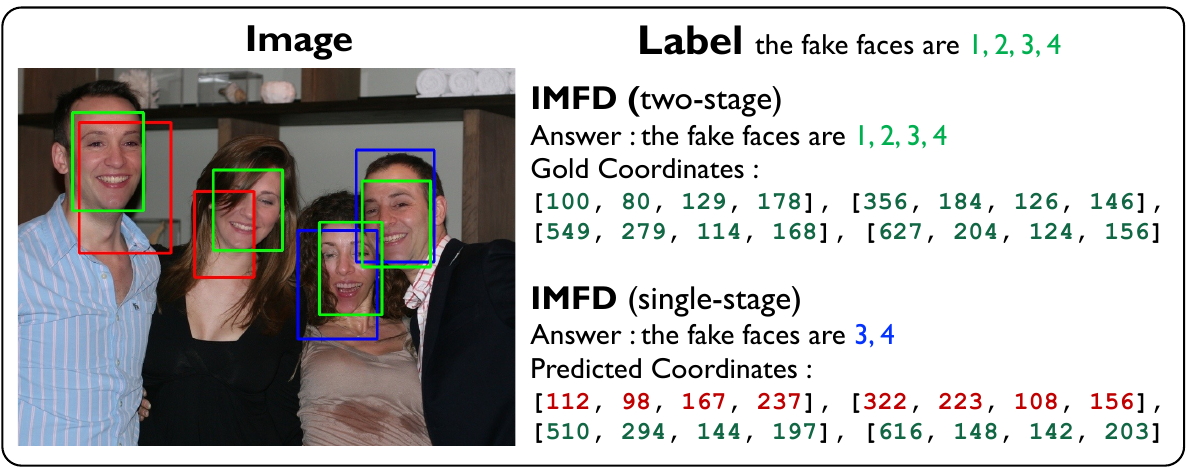}
    \caption{Face bounding-box prediction and forgery detection by IMFD in two-stage and single-stage settings.}
    \label{fig:bbox}
\end{figure}

\noindent \textbf{Error Analysis.}
We calculated COCO-style Average Precision (AP) for face bounding box prediction~\cite{lin2015microsoftcococommonobjects}. The AP score of 81.9 for the ALL dataset highlights the importance of high-quality face localization for the single-stage IMFD in Table~\ref{tab:all_results}. We also confirmed that our method can predict face bounding boxes for a large number of faces in a single image, achieving an AP of 83.6 for the LARGE dataset. Figure~\ref{fig:bbox} provides a qualitative comparison between the two-stage and single-stage settings. The two-stage IMFD accurately detects all fake faces using the ground-truth face coordinates, while the single-stage IMFD only identifies a subset of fake faces based on the predicted bounding boxes. This highlights that errors in bounding box prediction for single-stage IMFD can directly affect the accuracy of fake face detection.

\noindent \textbf{Impact of Auxiliary Details.}
We analyzed the effectiveness of auxiliary details in instructions for multi-face forgery detection. Table~\ref{tab:infpriming} presents the results obtained using IMFD. Even without auxiliary details, IMFD ($F1_c$) substantially outperforms the best two-stage baselines in Table~\ref{tab:all_results}.
Solely injecting image resolution was not effective. Resolution alone lacks spatial correspondence and thus cannot satisfy the left-to-right labeling required by instruction-based detection. A more detailed analysis of the effect of input image resolution is provided in Appendix~\ref{app:resolution_analysis}. Additional Grad-CAM case studies are provided in Appendix~\ref{app:case_study}.


\section{Conclusion and Discussion}
We proposed IMFD, a single-stage end-to-end instruction-based detector that jointly localizes faces and reasons about their authenticity using LVLMs. We convert existing multi-face forgery datasets into an instruction-based format with auxiliary details. On OpenForensics, IMFD consistently outperforms strong two-stage and single-stage baselines. In the future, we will further investigate various LVLM backbones, improve bounding-box quality, and extend the framework to video.

\section*{Limitations}

Our study has several limitations. First, experiments are limited to OpenForensics because other multi-face forgery benchmarks are not publicly available. Second, the single-stage setting is sensitive to face localization quality, and errors in predicted bounding boxes directly affect forgery detection performance. Third, we focus only on image-based forgery detection and do not address temporal consistency in videos. In addition, the relatively small number of zero-fake samples may limit robustness in such cases. Future work includes evaluation on more diverse benchmarks, improving box prediction quality, and extending the framework to video.

\section*{Acknowledgments}
This work was supported by Institute of Information \& communications Technology Planning \& Evaluation (IITP) grant funded by the Korea government(MSIT) (RS-2026-25569031, Development of an AIBOM-Based Platform for Strengthening the Reliability and Security of the AI Supply Chain).
This study was financially supported in part by the National Research Foundation of Korea (NRF) grant funded by the Korean government (MSIT) (RS-2026-25580044).
This work was supported by the Korea Basic Science Institute (KBSI) Young Researcher Infrastructure Support Program funded by the Ministry of Science and ICT (MSIT) (Grant No. RS-2026-25473607).
Following are results of a study on the “Convergence and Open Sharing System” Project, supported by the Ministry of Education and National Research Foundation of Korea.

\bibliography{custom}

\appendix

\section{Implementation Details}
\label{app:implementation_details}

We employed InternVL2-2B~\cite{chen2024internvl} as the backbone model and fine-tuned it with LoRA~\cite{hu2022lora}. Table~\ref{tab:model_settings} summarizes the LVLM backbones used in our experiments, including their parameter sizes and underlying LLMs. Experiments were conducted using the Transformers library with a random seed of 42 on an NVIDIA A6000 GPU. The input image resolution was fixed at 448$\times$448. We set $\tau_{\mathrm{iou}}$ and $\tau_{\mathrm{skip}}$ to 0.01 and 0.8, respectively. We set $\tau_{\mathrm{tp}}$ and $\tau_{\mathrm{fp}}$ to 0.0 and -2.0, respectively.

\begin{figure*}[t]
    \centering
    \includegraphics[width=1.0\linewidth]{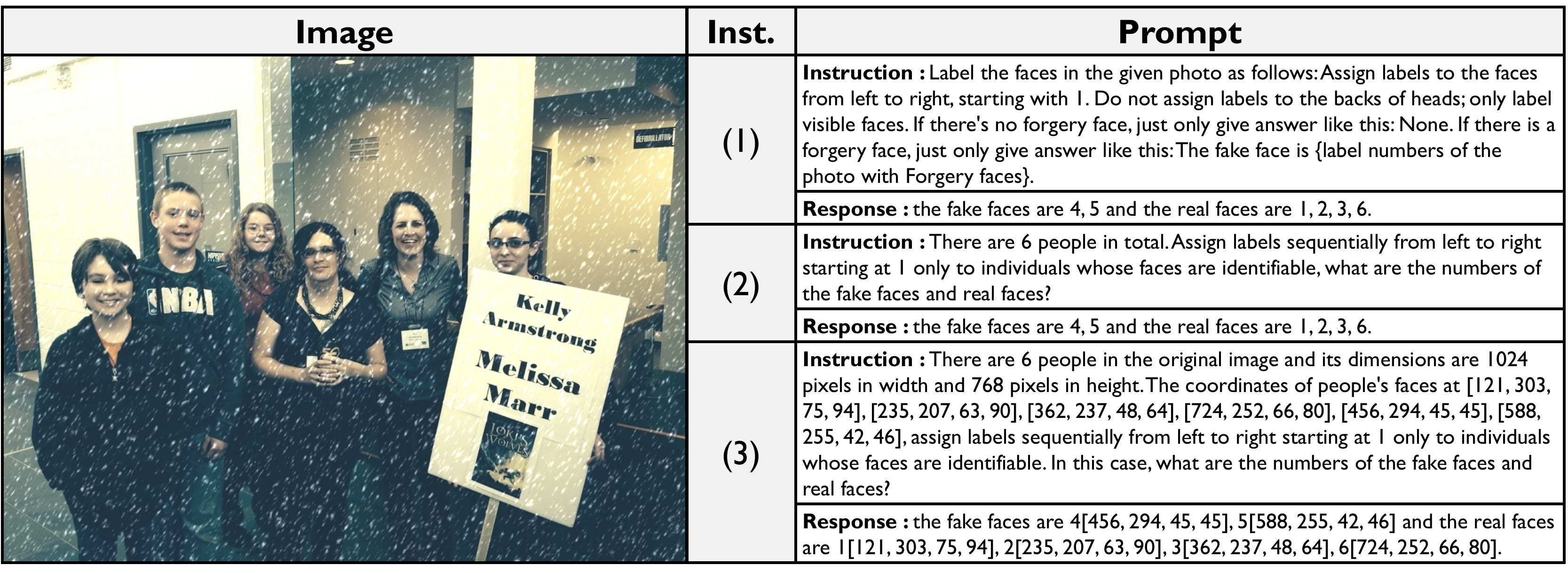}
    \caption{Face forgery detection instructions and responses.}
    \label{tab:inst_settings_2}
\end{figure*}

Figure~\ref{tab:inst_settings_2} shows the three instruction settings used in the two-stage IMFD analysis. Setting (1) uses the instruction without auxiliary details. Setting (2) adds only the total number of people. Setting (3) incorporates the total number of people, image resolution, and face coordinates. These settings are used in the Grad-CAM case study to analyze how auxiliary details affect instruction grounding and multi-face forgery detection.



\section{Upper-bound Analysis}
\label{app:upper_bound_analysis}

To analyze the upper bound of instruction-based multi-face forgery detection independently of face localization errors, we additionally evaluate several LVLM backbones using ground-truth face bounding-box coordinates injected into the instruction. This setting removes the effect of bounding-box prediction noise and allows us to compare instruction variants under controlled conditions. We consider three instruction variants: (1) the base instruction without auxiliary details, (2) the instruction augmented with only the total number of people, and (3) the instruction augmented with the total number of people, image resolution, and face coordinates.

Table~\ref{tab:upper_bound_results} shows that augmenting the instruction with auxiliary details consistently improves performance across LVLM backbones. In particular, instruction variant (3), which includes both global and spatial cues, achieves the best overall performance in most cases. This confirms that the benefit of auxiliary details is not limited to a specific model family. We also observe that increasing the LLM size does not always lead to better performance. For example, InternVL2-2B remains competitive with, and in some cases outperforms, larger variants such as InternVL2-4B and InternVL2-8B. These results suggest that performance in this task depends more on how effectively the model aligns visual faces with the instruction than on model scale alone.

\begin{table}[t!]
\renewcommand{\arraystretch}{0.95}
\centering
\resizebox{1.0\columnwidth}{!}{%
\begin{tabular}{ccccc|ccc}

\toprule
\rowcolor{gray!10}
& & \multicolumn{3}{c}{\textbf{ALL}} & \multicolumn{3}{c}{\textbf{LARGE}}  \\
\cline{3-8}
\rowcolor{gray!10}
\multirow{-2}{*}{\textbf{Model}}  & \multirow{-2}{*}{\textbf{Inst.}}  
    & \textbf{F1$_c$} & \textbf{F1$_a$} & \textbf{Acc}   
    & \textbf{F1$_c$} & \textbf{F1$_a$} & \textbf{Acc} \\

\noalign{\vskip -30pt}
\multirow{1}{*}{} &  &  &  &  &  &  &  \\
&  &  &  &  &  &  &  \\
&  &  &  &  &  &  &  \\
\noalign{\vskip -4pt}

\midrule

\multirow{3}{*}{\makecell{LLaVA-1.5\\-Vicuna}}   
    &   (1)   & \underline{0.77} &0.41 & \underline{0.56} & 0.84 & 0.52 & 0.43 \\   
    &   (2)   & 0.73 & \underline{0.56} & 0.32 & 0.93& \underline{\textbf{0.81}} & \textbf{0.64} \\ 
    &   (3)   & \textbf{0.91}$^\dagger$ & \textbf{0.92}$^\dagger$ & \textbf{0.78}$^\dagger$ & \textbf{0.94}& \textbf{0.81}$^\dagger$ & 0.55 \\ 
\midrule

\multirow{3}{*}{\makecell{LLaVA-1.6\\-Vicuna}}
    &   (1)   & \underline{0.86} &0.52& \underline{0.74} & 0.89 & 0.57 & 0.55\\
    &   (2)   & 0.73 & \underline{0.56} & 0.32 & \underline{0.93} & \underline{\textbf{0.81}} & \underline{0.64} \\ 
    &   (3)   & \textbf{0.96}$^\dagger$ & \textbf{0.94}$^\dagger$ & \textbf{0.90}$^\dagger$& \textbf{0.94}$^\dagger$ & \textbf{0.81}$^\dagger$ & \textbf{0.76}$^\dagger$ \\ 
\midrule

\multirow{3}{*}{\makecell{InternVL2\\-2B}} 
    &   (1)   & 0.91 & 0.61 & 0.83 & 0.94 & 0.65 & 0.74\\
    &   (2)   & \underline{0.94} & \underline{0.81} & \underline{0.87} & \underline{0.96} & \underline{0.90} & \underline{0.79} \\ 
    &   (3)   & \textbf{0.98}$^\dagger$ & \textbf{0.98}$^\dagger$ & \textbf{0.94}$^\dagger$ & \textbf{0.97}$^\dagger$ & \textbf{0.98}$^\dagger$ & \textbf{0.87}$^\dagger$ \\ 
\midrule

\multirow{3}{*}{\makecell{InternVL2\\-4B}}     
    &   (1)   & 0.91 & 0.66 & 0.84 & 0.94 & 0.77 & 0.73 \\
    &   (2)   & \underline{0.95} & \underline{0.78} & \underline{0.88} & 0.95 & \underline{0.78} & \underline{0.76} \\ 
    &   (3)   & \textbf{0.96}$^\dagger$ &  \textbf{0.91}$^\dagger$ & \textbf{0.89}$^\dagger$ & \textbf{0.95}& \textbf{0.96}$^\dagger$ & \textbf{0.80}$^\dagger$ \\ 
\midrule

\multirow{3}{*}{\makecell{InternVL2\\-8B}}       
    &   (1)   & 0.91 & 0.65 & 0.84 & 0.94 & 0.76 & 0.76\\
    &   (2)   & \underline{0.94} & \underline{0.77}& \underline{0.86} & \textbf{0.95} & \underline{0.90}& \underline{0.77} \\ 
    &   (3)   & \textbf{0.96}$^\dagger$ & \textbf{0.98}$^\dagger$& \textbf{0.90}$^\dagger$ & \textbf{0.95} & \textbf{0.97}$^\dagger$&\textbf{0.80}$^\dagger$ \\ 
\midrule

\multirow{3}{*}{\makecell{MiniCPM\\-Llama3-V-v2.5}}
    &   (1)   & 0.53 & 0.15 & 0.25 & 0.50 & 0.18 & 0.00\\ 
    &   (2)   & \underline{0.73} & \underline{0.56} & \underline{0.32}  & 0.92 & \underline{0.79} &  \underline{0.59} \\ 
    &   (3)   & \textbf{0.90}$^\dagger$ & \textbf{0.95}$^\dagger$ & \textbf{0.77}$^\dagger$ & \textbf{0.95} & \textbf{0.97}$^\dagger$ & \textbf{0.80}$^\dagger$  \\
\midrule

\multirow{3}{*}{\makecell{MiniCPM\\-V-v2.6}}
    &   (1)   & 0.49 & 0.11 & 0.26 & 0.43 & 0.13 &  0.00\\
    &   (2)   & \underline{0.72} & \underline{0.46} & \underline{0.29} & \textbf{0.93} & \underline{0.81} & \underline{0.64} \\ 
    &   (3)   & \textbf{0.91}$^\dagger$ & \textbf{0.96}$^\dagger$ & \textbf{0.77}$^\dagger$ & 0.92 & \textbf{0.95}$^\dagger$ & \textbf{0.70}$^\dagger$ \\ 
\bottomrule
\end{tabular}
}

\caption{Upper-bound results on \textbf{OpenForensics} obtained by injecting ground-truth face bounding-box coordinates into the instruction. (1), (2), and (3) denote the three instruction variants described in the main text. \textbf{ALL} and \textbf{LARGE} denote evaluation on the full test set and on the subset of images containing more than the average number of faces (2.63), respectively. $\dagger$ indicates that the improvement is significant (\textit{p}$<$0.01) compared with the underlined score, as determined by paired bootstrap resampling~\cite{koehn-sig}.}
\label{tab:upper_bound_results}
\end{table}

\section{Impact of Input Image Resolution}
\label{app:resolution_analysis}

To provide a more detailed analysis of image resolution, we evaluate IMFD under different input resolutions using InternVL2-2B as the backbone and instructions augmented with ground-truth face coordinates. Table~\ref{tab:imagesize} reports the results on ALL and LARGE. 
As the input resolution increases from 112$\times$112 to 672$\times$672, the performance improves consistently across all evaluation metrics. 

In particular, on ALL, the F1$_c$ and F1$_a$ scores reach 0.981 and 0.985, respectively, at 672$\times$672. A similar trend is observed on LARGE. These results suggest that higher-resolution inputs provide more discriminative visual details for instruction-based multi-face forgery detection.

\begin{table}[t]
\renewcommand{\arraystretch}{1}
\centering
\resizebox{1.0\columnwidth}{!}{
\begin{tabular}{ccccccc}
    \toprule
    \rowcolor{gray!10}
    & \multicolumn{3}{c}{\textbf{ALL}} & \multicolumn{3}{c}{\textbf{LARGE}} \\
    \rowcolor{gray!10}
    \textbf{Image Rel.} & \textbf{F1$_c$} & \textbf{F1$_a$} & \textbf{Acc} & \textbf{F1$_c$} & \textbf{F1$_a$} & \textbf{Acc} \\
    \midrule
    112x112 & 0.9173 & 0.9414 & 0.6538 & 0.8894 & 0.9401 & 0.6538 \\
    224x224 & 0.9414 & 0.9492 & 0.7424 & 0.9260 & 0.9566 & 0.7424 \\
    448x448 & 0.9762 & 0.9846 & 0.9359 & 0.9744 & 0.9838 & 0.8739 \\
    672x672 & \textbf{0.9810} & \textbf{0.9849} & \textbf{0.9484} & \textbf{0.9808} & \textbf{0.9841} & \textbf{0.9008} \\
    \bottomrule
\end{tabular}
}
\caption{Impact of input image resolution on IMFD using InternVL2-2B.}
\label{tab:imagesize}
\end{table}

\section{Grad-CAM Case Study}
\label{app:case_study}

\begin{figure*}[t]
    \centering
    \includegraphics[width=0.8\linewidth]{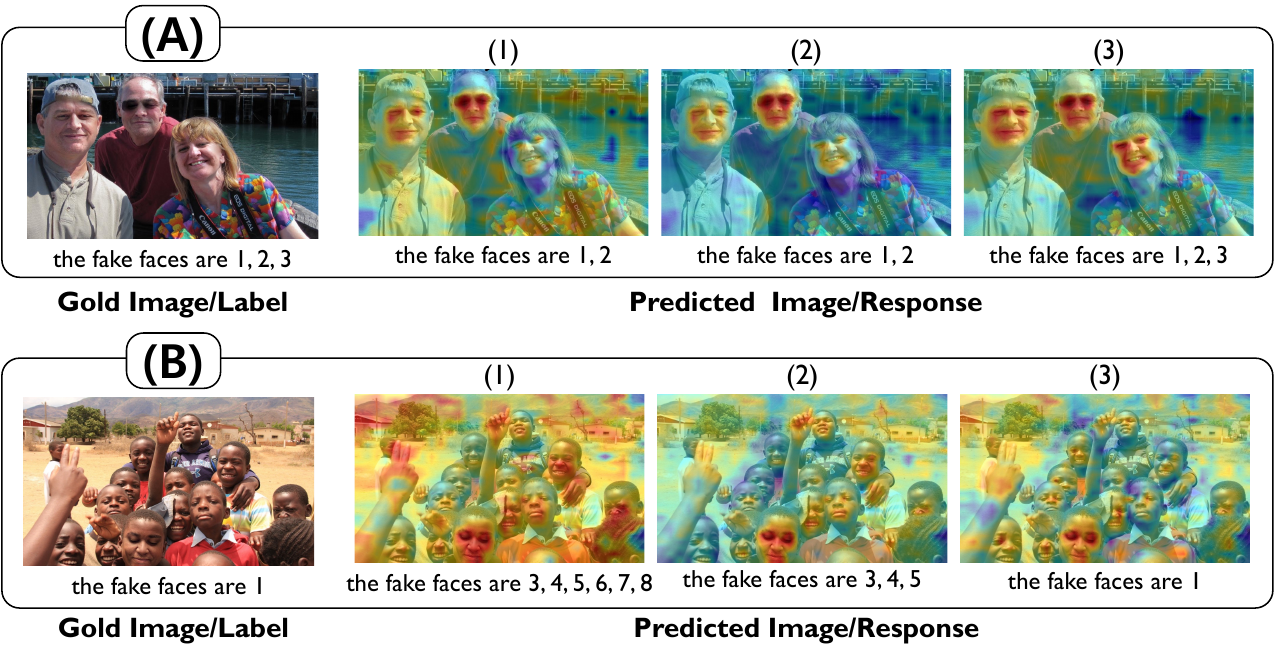}
    \caption{Original and Grad-CAM visualizations for the 18th layer on images containing forged faces.}
    \label{fig:grad}
\end{figure*}

We evaluated whether instruction-based LVLMs can parse multi-face images with left-to-right numeric labels by visualizing IMFD Grad-CAM heat maps on test samples~\cite{8237336}. Figure~\ref{fig:grad} shows the results.
(1) uses the original instructions without auxiliary details, (2) adds only the total number of people, and (3) incorporates the number of people, image resolution, and face coordinates.
In example (A), the model's attention aligns with its predictions: with (1) and (2), it focuses on and labels only the first two forged faces, while with (3), attention extends to all three faces. Example (B) further shows that, when using (3), the LVLM with auxiliary details can localize and detect fake faces even in images with many faces.

\begin{table*}[t]
\centering
\renewcommand{\arraystretch}{1.05}
\resizebox{\textwidth}{!}{
\begin{tabular}{lcll}
    \toprule
    \rowcolor{gray!10}
    \textbf{Model} & \textbf{Params} & \textbf{Backbone LLM} & \textbf{HuggingFace Repository} \\
    \midrule
    \texttt{LLaVA-1.5} & 7B & vicuna-7b-v1.5 & \url{https://huggingface.co/liuhaotian/llava-v1.5-7b} \\
    \texttt{LLaVA-1.6} & 7B & vicuna-7b-v1.5 & \url{https://huggingface.co/liuhaotian/llava-v1.6-vicuna-7b} \\
    \texttt{InternVL2-2B} & 2.2B & internlm2-chat-1\_8b & \url{https://huggingface.co/OpenGVLab/InternVL2-2B} \\
    \texttt{InternVL2-4B} & 4.2B & Phi-3-mini-128k-instruct & \url{https://huggingface.co/OpenGVLab/InternVL2-4B} \\
    \texttt{InternVL2-8B} & 8.1B & internlm2\_5-7b-chat & \url{https://huggingface.co/OpenGVLab/InternVL2-8B} \\
    \texttt{MiniCPM-Llama3-V-2.5} & 8.1B & Llama3-8B-Instruct & \url{https://huggingface.co/openbmb/MiniCPM-Llama3-V-2_5} \\
    \texttt{MiniCPM-V-2.6} & 9.5B & Qwen2-7B-Instruct & \url{https://huggingface.co/openbmb/MiniCPM-V-2_6} \\
    \bottomrule
\end{tabular}
}
\caption{LVLM backbones used in this study, including parameter sizes, backbone LLMs, and model repositories for reproducibility.}
\label{tab:model_settings}
\end{table*}

\end{document}